\documentclass[11pt, a4paper, logo, copyright, onecolumn, nonumbering]{googledeepmind}

\pdftrailerid{redacted}

\usepackage{kantlipsum, lipsum}
\usepackage{dsfont}
\usepackage{multirow}
\usepackage{array}
\usepackage{amssymb} %
\usepackage{pifont} %
\usepackage{makecell}
\usepackage{algorithm}
\usepackage{algpseudocode}
\usepackage{amsmath}
\usepackage{multicol}
\usepackage{float}
\usepackage{siunitx}
\usepackage{tabularx}
\usepackage{booktabs}

\usepackage{microtype}
\usepackage[pagebackref, breaklinks, colorlinks]{hyperref}    %
\usepackage{url}            %
\usepackage{amsfonts}       %
\usepackage{nicefrac}       %
\usepackage{xcolor}         %
\usepackage{float}
\usepackage{multirow}
\usepackage{multicol}
\usepackage{colortbl}
\usepackage{soul}
\usepackage{caption}
\usepackage{amsmath} 
\usepackage{array} 
\usepackage{tcolorbox}
\usepackage{pifont}
\usepackage{placeins}
\usepackage{enumitem}
\usepackage{xspace}
\usepackage{makecell}
\usepackage{wrapfig}
\usepackage{amssymb}
\usepackage{adjustbox}
\usepackage{xcolor}
\definecolor{GoogleBlue}{HTML}{1A73E8}
\definecolor{GoogleBlueLight}{HTML}{e8f0fe}
\definecolor{GoogleBlueDark}{HTML}{174ea6}
\definecolor{GoogleRed}{HTML}{D93025}
\definecolor{GoogleGray}{HTML}{767676}
\definecolor{figtextcolor}{HTML}{333333}
\definecolor{perceptioncol}{HTML}{1A73E8}
\definecolor{modelingcol}{HTML}{925BCF}
\definecolor{manipulationcol}{HTML}{C440A7}
\definecolor{reasoningcol}{HTML}{DA2F77}

\usepackage{xurl}

\usepackage{graphicx}
\graphicspath{{figures/}}

\title{EgoReasoner: Learning Egocentric 4D Reasoning via Task-Adaptive Structured Thinking} 
\usepackage[numbers, sort&compress]{natbib}
\correspondingauthor{
  zhu.fang@northeastern.edu
}
\definecolor{googleblue}{HTML}{4285F4}
\definecolor{googlered}{HTML}{EA4335}
\definecolor{googlegreen}{HTML}{34A853}
\definecolor{googleyellow}{HTML}{FBBC04}
\reportnumber{} %
\usepackage{amsmath}
\usepackage{amssymb}
\usepackage{mathtools}
\usepackage{amsthm}
\usepackage{bm}
\usepackage{booktabs}
\usepackage{multirow}
\usepackage{tcolorbox} 
\tcbuselibrary{most}   %
\usepackage{hyperref}
\usepackage{cleveref}  %
\usepackage{float} 
\author{
Fangrui Zhu\textsuperscript{1,*},
Yunfeng~Xi\textsuperscript{2},
Jianmo~Ni\textsuperscript{3},
Mu~Cai\textsuperscript{3},
Boqing~Gong\textsuperscript{3},
Long~Zhao\textsuperscript{3},
Chen~Qu\textsuperscript{2},
Ian~Miao\textsuperscript{2},
Yi~Li\textsuperscript{2},
Cheng~Zhong\textsuperscript{2},
Huaizu~Jiang\textsuperscript{1,$\bm{\dagger}$},
Shwetak~Patel\textsuperscript{2,$\bm{\dagger}$} \\
\vspace{0.5em}
\normalfont\fontsize{8}{10}\selectfont
\textsuperscript{1}Northeastern University, 
\textsuperscript{2}Google Research,
\textsuperscript{3}Google DeepMind \\
\textsuperscript{$\dagger$}Co-last authors, 
\textsuperscript{*}Work done during an internship at Google
}

\usepackage{kantlipsum, lipsum}
\usepackage{dsfont}
\usepackage{multirow}
\usepackage{array}
\usepackage{amssymb} %
\usepackage{pifont} %
\usepackage{makecell}
\usepackage{algorithm}
\usepackage{algpseudocode}
\usepackage{amsmath}
\usepackage{multicol}
\usepackage{float}
\usepackage{siunitx}
\usepackage{tabularx}
\usepackage{booktabs}

\usepackage{microtype}
\usepackage[pagebackref, breaklinks, colorlinks]{hyperref}    %
\usepackage{url}            %
\usepackage{amsfonts}       %
\usepackage{nicefrac}       %
\usepackage{xcolor}         %
\usepackage{float}
\usepackage{multirow}
\usepackage{multicol}
\usepackage{colortbl}
\usepackage{soul}
\usepackage{caption}
\usepackage{amsmath} 
\usepackage{array} 
\usepackage{tcolorbox}
\usepackage{pifont}
\usepackage{placeins}
\usepackage{enumitem}
\usepackage{xspace}
\usepackage{makecell}
\usepackage{wrapfig}
\usepackage{amssymb}
\usepackage{adjustbox}
\usepackage{xcolor}
\definecolor{GoogleBlue}{HTML}{1A73E8}
\definecolor{GoogleBlueLight}{HTML}{e8f0fe}
\definecolor{GoogleBlueDark}{HTML}{174ea6}
\definecolor{GoogleRed}{HTML}{D93025}
\definecolor{GoogleGray}{HTML}{767676}
\definecolor{figtextcolor}{HTML}{333333}
\definecolor{perceptioncol}{HTML}{1A73E8}
\definecolor{modelingcol}{HTML}{925BCF}
\definecolor{manipulationcol}{HTML}{C440A7}
\definecolor{reasoningcol}{HTML}{DA2F77}

\usepackage{xurl}

\usepackage{graphicx}
\graphicspath{{figures/}}

\def\eg{\textit{e.g.}\@\xspace}

\definecolor{mygreen}{RGB}{1,113,0}

\definecolor{myyellow}{RGB}{243, 210, 147}
\definecolor{myred}{RGB}{236, 131, 126}
\newcommand{\hlyellow}[1]{{\sethlcolor{myyellow}\hl{#1}}}
\newcommand{\hlred}[1]{{\sethlcolor{myred}\hl{#1}}}

\newcommand{\rowA}{\rowcolor{blue!6}}
\newcommand{\rowB}{\rowcolor{gray!20}}

\newcolumntype{R}[2]{%
    >{\adjustbox{angle=#1,lap=\settowidth{\width}{#2},caption={#2}}\bgroup}%
    l%
    <{\egroup}%
}

\newcommand{\model}{\texttt{EgoReasoner}}

\begin{abstract}
Egocentric video understanding is inherently complex due to the dynamic 4D nature of the environment, where camera motion and object displacements necessitate a continuous re-evaluation of spatial relations. 
In this work, we target a suite of under-explored egocentric 4D reasoning tasks, including fixture interaction counting, viewpoint-relative fixture location, object movement itinerary tracking, and stationary object localization, that require fundamentally different cognitive operations: spatial anchoring, temporal tracking, and duration reasoning.
We observe that these structural differences make task-agnostic approaches insufficient: generic Chain-of-Thought methods lack task-appropriate reasoning primitives, and uniform reinforcement learning actively destabilizes performance on spatial tasks.
To address this, we propose \model{}, a two-stage framework that aligns both the reasoning scaffold and the reward signal to each task's cognitive structure.
In the first stage, Task-Adaptive Thinking Templates guide the synthesis of structured CoT traces that teach the model to reason adaptively across task types via supervised fine-tuning.
In the second stage, task-aware reward functions verify entity grounding, temporal alignment, and task-adaptive logical consistency, selectively strengthening each reasoning pathway via reinforcement fine-tuning with GRPO.
Our 3B-parameter model, trained on only 16K samples, achieves 37.5\% average accuracy on the challenging HD-EPIC benchmark, surpassing Qwen2.5-VL-7B (25.7\%) by over 10 points.
\end{abstract}

\begin{document}

\newpage
\maketitle

\begin{figure}[H]
    \includegraphics[width=\textwidth]{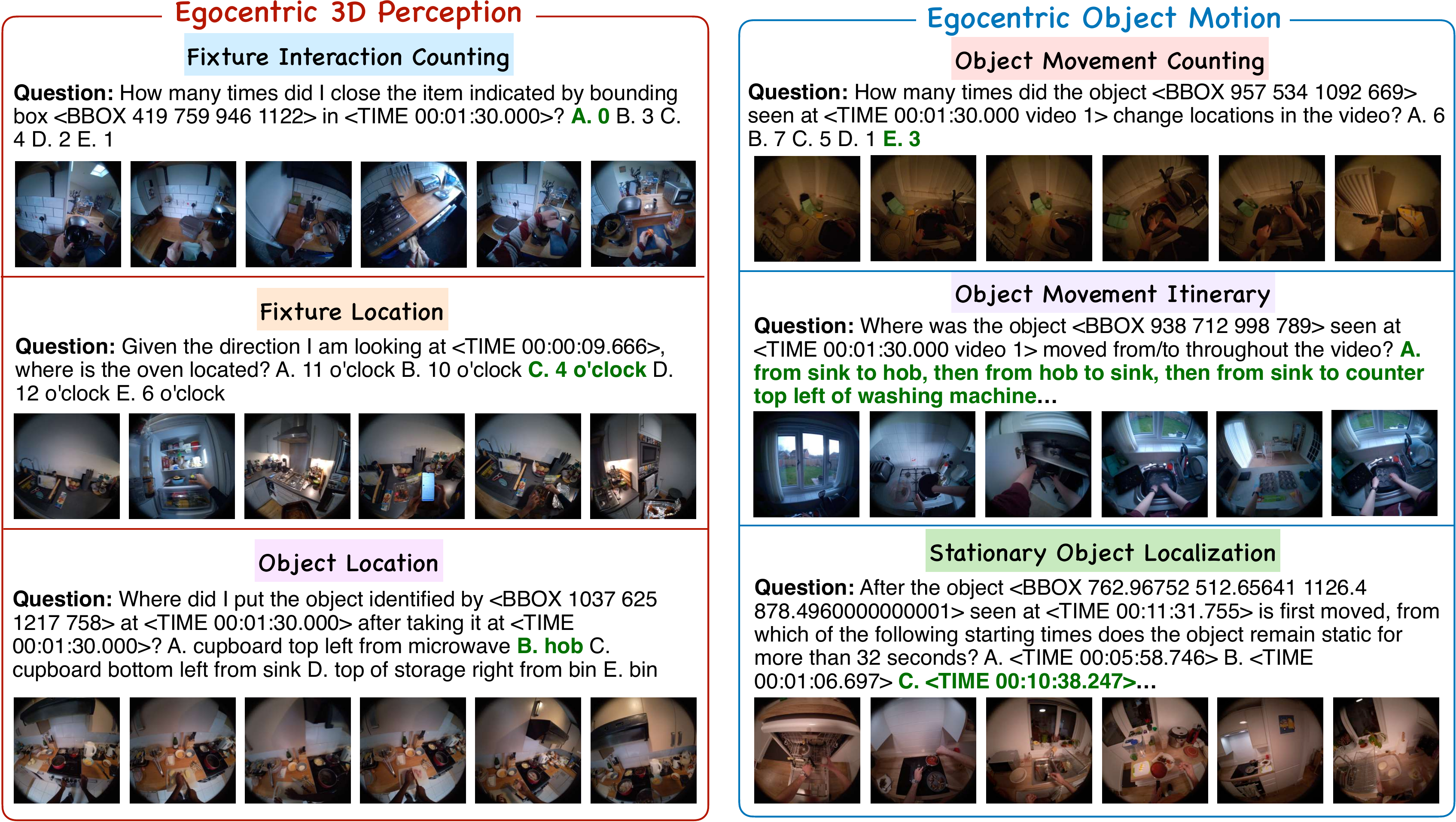}
    \caption{\label{fig:task}\textbf{Illustration of 4D Reasoning tasks.}  We distinguish between fixtures~(static objects in the scene) and objects~(moving entities) to evaluate complex spatial-temporal understanding~(from \cite{perrett2025hd}). For each task, we display six frames sampled from the original video. These tasks require the model to reason about fixture locations using ``clock-face'' orientations, track object movement itineraries, and count interactions despite constant ego-motion. Correct answers are indicated in \textbf{\textcolor{mygreen}{green bolded}} text. Some options are omitted due to space limit.}
\end{figure}
\section{\label{sec:intro}Introduction}
\begin{figure}[t]
    \includegraphics[width=\textwidth]{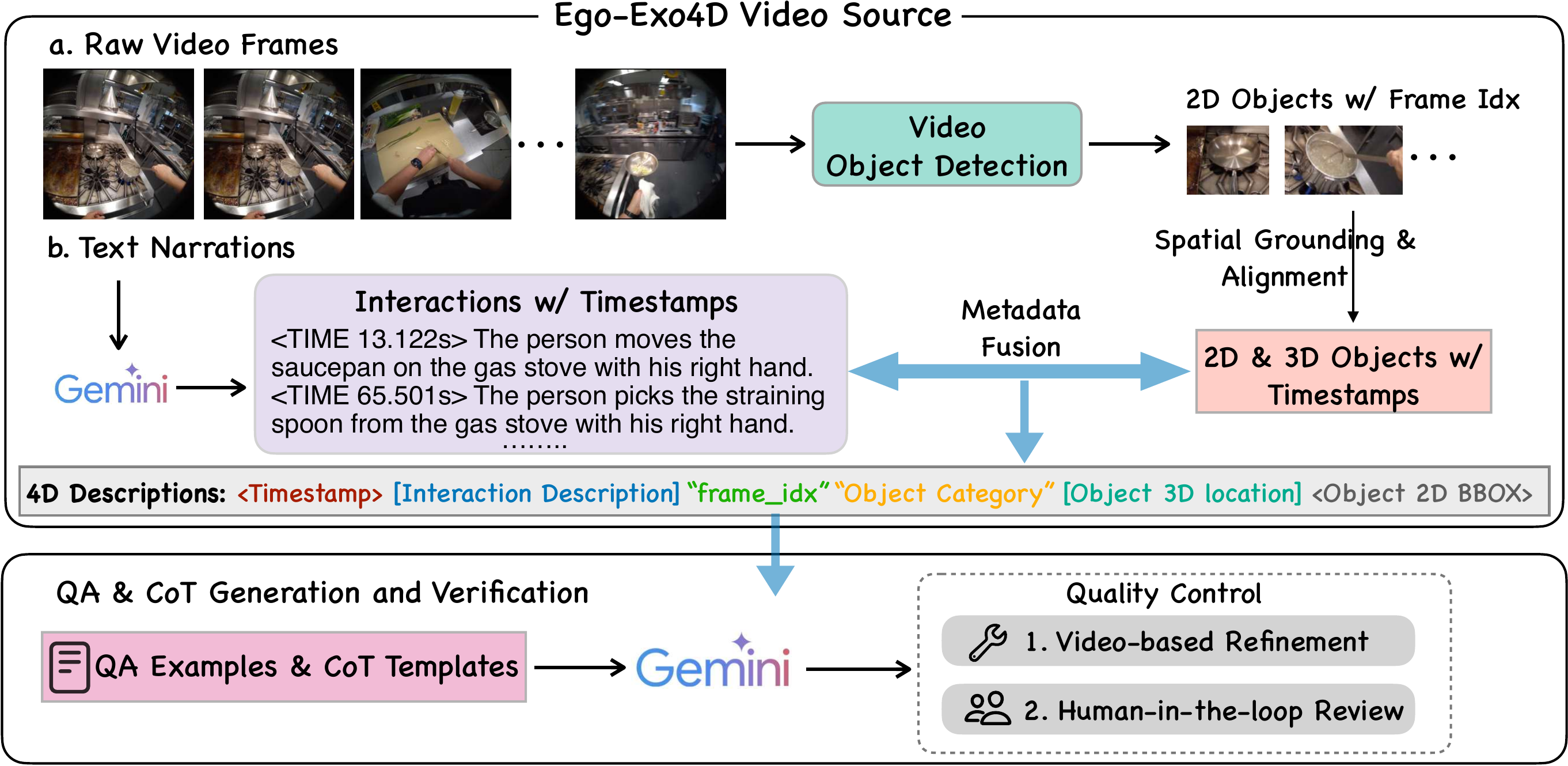}
    \caption{\label{fig:data}\textbf{Automated Metadata-Driven Pipeline for QA and CoT Generation.} We first preprocess the video data by (a) extracting precise 2D \& 3D object trajectories via video object detection and SLAM-based camera alignment; and (b) merging these with Gemini-refined text narrations to create unified 4D Descriptions. 
    These descriptions are processed through Task-Adaptive Thinking Templates to generate grounded QA pairs and CoT traces.}
\end{figure}
\begin{figure}[t]
    \includegraphics[width=\textwidth]{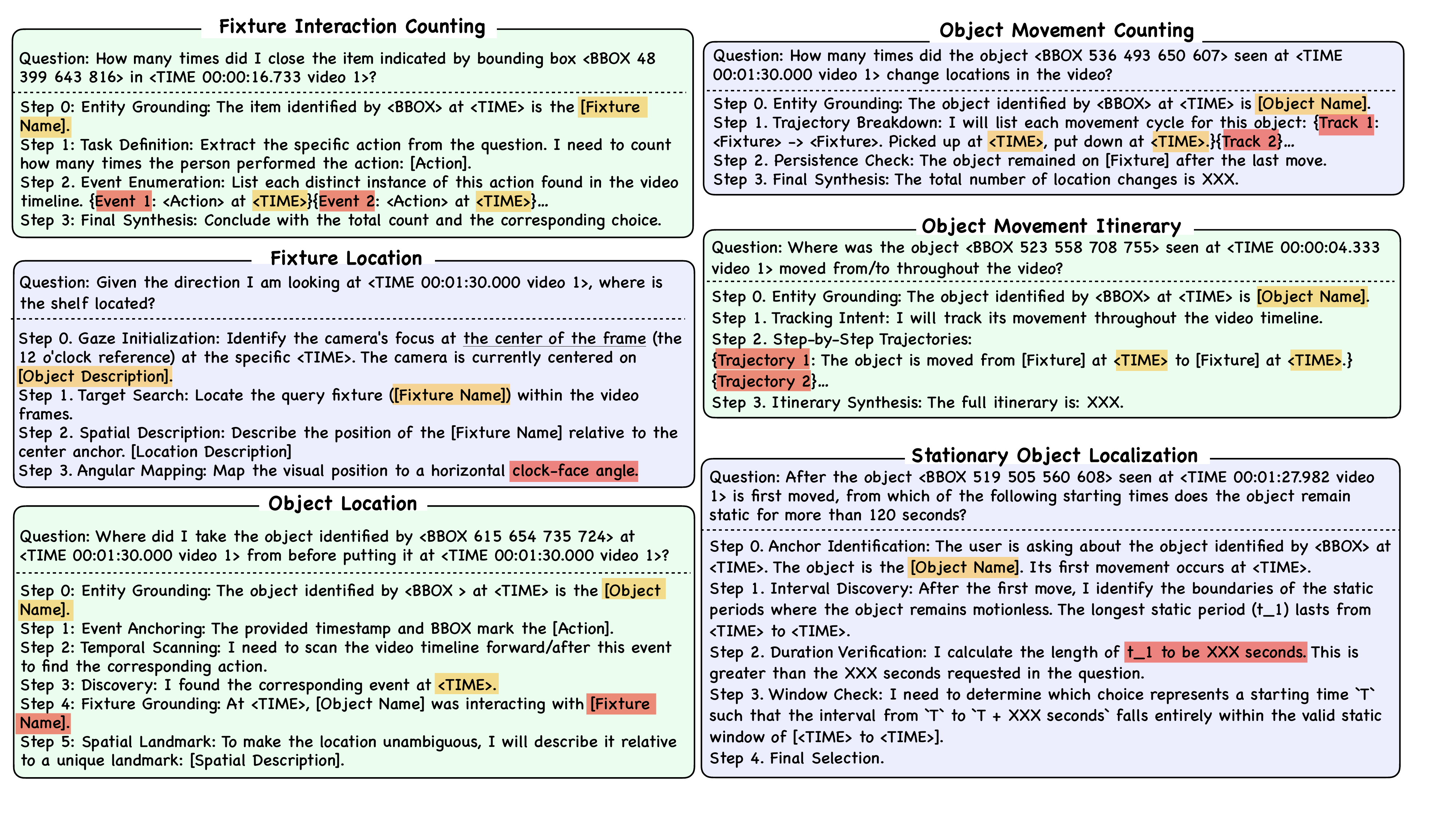}
    \caption{\label{fig:think_temp}\textbf{Task-Adaptive Thinking Templates.} Our templates decompose 4D reasoning into structured sub-steps. \hlyellow{Yellow highlights} indicate grounded entity names (fixtures and objects) / timestamps, while \hlred{red highlights} denote specific spatial-temporal metadata, such as angular orientations, and trajectory segments.}
\end{figure}
\begin{figure}[t]
    \includegraphics[width=\textwidth]{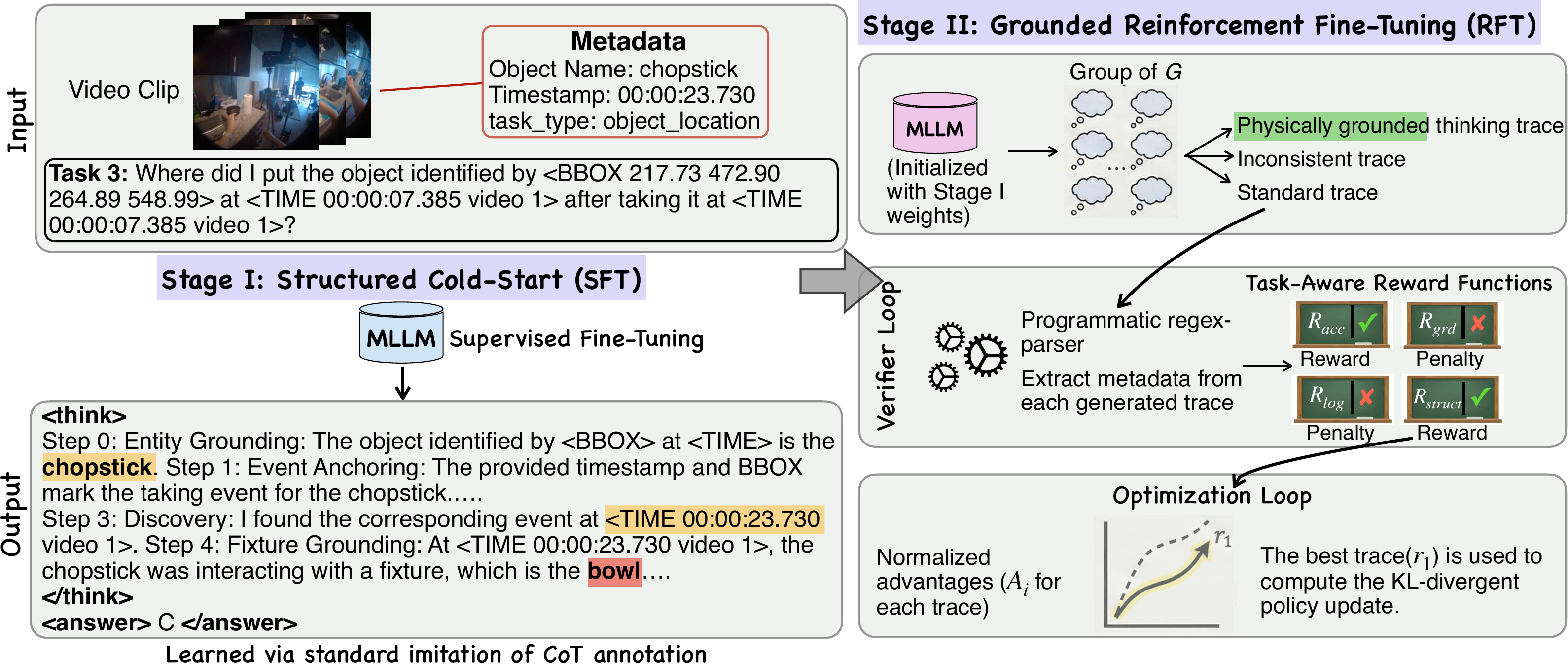}
    \caption{\label{fig:train_pipeline}\textbf{Overview of the Training Paradigm of \model{}.} Stage I Structured Cold-Start (SFT): The MLLM is fine-tuned to imitate structured reasoning traces. It learns to generate template-based \texttt{<think>} blocks that anchor \hlyellow{entities/timestamps} and identify \hlred{fixture} prior to providing the final \texttt{<answer>}. Stage II: Grounded Reinforcement Fine-Tuning (RFT): The model is optimized via GRPO to ensure physical verifiability.}
\end{figure}
Egocentric video understanding is a cornerstone of embodied AI and augmented reality, providing the sensory foundation for agents to interact naturally with the human world~\cite{astra, aria}. 
Unlike third-person videos, the first-person perspective necessitates interpreting human intent and tracking complex object dynamics from a highly non-stationary viewpoint, where both the scene layout and the observer's reference frame change continuously.
This domain serves as a rigorous testbed for \emph{4D spatial-temporal reasoning}, the ability to maintain a consistent world model of both static ``fixtures'' (such as an oven or a sink) and dynamic ``objects'' (such as a pot being moved) as they interact over time, even as the camera's ego-motion constantly reshapes the visual evidence.

In this work, we target a suite of challenging, under-explored tasks (\cref{fig:task}) that expose the unique demands of egocentric 4D reasoning: fixture interaction counting, fixture location via clock-face orientations, object location, object movement counting, object movement itinerary tracking, and long-term stationary object localization.
These tasks are uniquely featured in HD-EPIC~\cite{perrett2025hd}.
Collectively, these tasks require three fundamentally different cognitive operations: spatial anchoring relative to a moving camera, temporal tracking of object trajectories across long horizons, and duration reasoning over static intervals. 
We identify three core challenges where existing methods fall short.

First, egocentric spatial reasoning requires anchoring to a moving camera.
Determining that ``the oven is at 4 o'clock'' requires computing an angular offset relative to the camera's gaze at a specific timestamp, a geometric operation that standard video MLLMs~\cite{cheng2024videollama,zhang2024long,zhang2024video,feng2025video} have no mechanism for.
Existing Chain-of-Thought~(CoT) methods~\cite{shao2024visual, ma2025spatialreasoner, zhu2025struct2d} cannot express this operation either, since their generic reasoning traces lack egocentric spatial priors such as clock-face mapping. 
Solving this task requires decomposing spatial reasoning into reference-frame identification and angular mapping, a task-adaptive structure that no existing approach provides.

Second, tracking object trajectories demands structured temporal bookkeeping over long horizons.
Reconstructing an itinerary like ``cupboard → counter → hob → sink'' requires maintaining a chronologically ordered log of object-fixture interactions across 3–5 minute videos, correctly associating the same object despite scale changes and hand occlusions. 
Prior video MLLMs~\cite{cheng2024videollama,zhang2024long,zhang2024video,feng2025video,wang2024qwen2} process such videos holistically and cannot produce timestamp-anchored trajectory logs.
Even reinforcement learning-enhanced methods like EgoThinker~\cite{pei2025egothinker} and VideoChat-R1~\cite{li2025videochat} use coarse IoU-based rewards that measure spatial or temporal overlap without verifying whether the model's reasoning correctly identifies the involved entities or respects temporal event boundaries. 
ST-Think~\cite{wu2026st}, which also targets egocentric 4D reasoning with GRPO, applies a uniform reverse-thinking strategy across all task types rather than adapting the reasoning structure to different task's requirements. 
What is needed instead is entity-level and timestamp-level verification to ensure that each reasoning step is anchored in the correct physical metadata.

Third, and most critically, each task demands a \emph{distinct reasoning structure}. 
Counting fixture interactions requires enumerating discrete events; itinerary tracking requires building a sequential log of locations; stationary object localization requires computing durations between temporal boundaries.
These are not minor variations on one reasoning pattern; they are structurally different cognitive operations that require different intermediate steps, different output formats, and different notions of correctness.
Yet generic CoT approaches~\cite{fei2024video, xu2025llava,zhang2023multimodal,cheng2024compressed,cheng2024spatialrgpt} apply the same reasoning pipeline regardless of task type, and prior reinforcement learning (RL) methods~\cite{guo2025deepseek,ouyang2025spacer, feng2025video, pei2025egothinker,li2025videochat,wang2025videorft} use same reward functions across all tasks. 
We show empirically (\cref{sec:experiment}) that task-agnostic RL not only underperforms but \emph{actively destabilizes} spatio-temporal reasoning: accuracy on fixture interaction counting and fixture location declines as training progresses, because uniform optimization pressure cannot accommodate the conflicting demands of structurally different tasks.

To address these challenges, we propose \model{}, a two-stage framework, where our key insight is that egocentric 4D reasoning benefits from aligning both the reasoning scaffold and the reward signal to each task's cognitive structure.
In the first stage, \emph{task-adaptive} templates guide the synthesis of structured CoT traces that, through supervised fine-tuning~(SFT), teach the student model to reason adaptively across task types.
In the second stage, reinforcement fine-tuning~(RFT) with \emph{task-aware} rewards selectively strengthens each reasoning pathway by verifying the intermediate outputs that each task's template is designed to produce.
Both stages are supported by an automated metadata-driven pipeline (\cref{fig:data}) that fuses 2D/3D object detections from SLAM-calibrated cameras with refined text narrations, producing dense 4D Descriptions that serve as the source for CoT synthesis \emph{and} as ground truth for reward computation.

\noindent \textbf{Stage I: Structured Cold-Start via Thinking Templates.}
We design Task-Adaptive Thinking Templates (\cref{fig:think_temp}) that decompose each 4D task into a sequence of grounded sub-steps whose structure matches the task's cognitive requirements: angular reasoning for spatial tasks, sequential logging for tracking tasks, event enumeration for counting tasks.
A strong teacher model~\cite{comanici2025gemini} generates structured CoT reasoning traces from ground-truth 4D metadata, and the student model (Qwen2.5-VL~\cite{wang2024qwen2}) is fine-tuned to imitate these traces, learning to reason adaptively across different task types.
This cold-start phase teaches the student the required logical format and spatial-temporal priors, providing a stable foundation for subsequent reinforcement learning.

\noindent \textbf{Stage II: Grounded Reinforcement Fine-Tuning.}
While Stage I establishes the reasoning format, supervised imitation alone cannot guarantee that the model's internal reasoning is factually grounded in the video.
Specifically, the entity names, timestamps, and spatial values produced during reasoning may not correspond to the physical metadata of the scene.
We further optimize the student model using GRPO~\cite{shao2024deepseekmath} with task-aware reward functions that verify intermediate reasoning steps against ground-truth 4D metadata.
We design three complementary reward signals: an \emph{entity grounding} reward that verifies correct object/fixture identification, a \emph{temporal grounding} reward that checks timestamp alignment within a soft-matching window, a \emph{logic} reward that enforces task-specific consistency (\eg, trajectory segment counts or clock-face angular accuracy).
Together, these rewards ensure that the model's reasoning traces are not merely well-formatted but \emph{grounded} in the physical reality of the 4D environment.

Our main contributions are as follows:
\begin{itemize}[leftmargin=*, nosep]
\item \textbf{Task-Adaptive Thinking Templates:} We design structured templates that teach MLLMs to reason adaptively across six egocentric 4D tasks, each decomposing the task into grounded sub-steps matched to its cognitive requirements. 
This enables a single model to handle tasks as diverse as angular spatial reasoning and long-horizon trajectory tracking.
\item \textbf{Task-Aware Reinforcement Learning:} We introduce fine-grained reward functions for GRPO that selectively strengthen each reasoning pathway by verifying entity grounding, temporal alignment, and task-adaptive logic against physical metadata. 
This provides per-step supervision that outcome-only or IoU-based rewards cannot offer.
\item \textbf{Strong Performance with a Small Model:} Our unified model achieves 37.5\% average accuracy on HD-EPIC, surpassing Qwen2.5-VL-7B (25.7\%) by over 10 points, with 59.5\% on Object Movement Counting (+26.5\% over the best baseline). 
\end{itemize}

\section{Related Work}
\subsection{Egocentric Video Reasoning.}
MLLMs have shifted the focus of egocentric video analysis from simple action recognition~\cite{kazakos2019epic,plizzari2022e2,sudhakaran2019lsta,wang2020symbiotic} to complex reasoning~\cite{ye2024mm,zhang2025exo2ego,pei2025egothinker,tian2025ego}, driven by benchmarks such as Ego4D~\cite{grauman2022ego4d}, Ego-Exo4D~\cite{grauman2024ego}, EgoSchema~\cite{mangalam2023egoschema}, and HD-EPIC~\cite{perrett2025hd} that demand long-horizon spatial-temporal understanding.
Among these, HD-EPIC uniquely formulates tasks in 4D space with 3D-grounded annotations via digital twins, requiring models to reason about fixture locations, object trajectories, and interaction counts, which go well beyond standard video QA. 
Recent egocentric MLLMs attempt to bridge general video understanding and first-person perspectives. Exo2Ego~\cite{zhang2025exo2ego} transfers exocentric knowledge via progressive multi-stage training, EgoThinker~\cite{pei2025egothinker} introduces spatio-temporal CoT with IoU-based RL rewards, Ego-R1~\cite{tian2025ego} applies chain-of-tool-thought for ultra-long egocentric videos, and MM-Ego~\cite{ye2024mm} proposes memory pointer prompting for long-horizon QA. 
However, these methods rely on visual captions, hand-object grounding, or tool-augmented retrieval without leveraging the rich sensor-level 3D metadata available in modern egocentric datasets. 
In contrast, our work synthesizes 4D reasoning traces directly from SLAM-calibrated metadata, providing higher-fidelity supervision that bypasses purely visual heuristics.

\subsection{Structured Thinking in MLLMs.}
Chain-of-Thought prompting~\cite{wei2022chain} and its multimodal extension~\cite{zhang2023multimodal} has become standard for eliciting step-by-step reasoning. 
In the visual domain, LLaVA-CoT~\cite{xu2025llava} introduces multi-stage tagged reasoning, while Video-of-Thought~\cite{fei2024video} and Chain-of-Frames~\cite{ghazanfari2025chain} extend CoT to video with frame-grounded traces. 
However, recent studies\cite{wang2025think,li2025think} show that unconstrained reasoning often degrades performance. Existing video-CoT frameworks~\cite{feng2025video, ma2025spatialreasoner, zhu2025struct2d, pei2025egothinker} produce generic traces lacking egocentric spatial-temporal priors. 
We instead introduce task-adaptive CoT templates with distinct sub-step sequences per task, supervised by ground-truth 4D metadata to ensure verifiably grounded reasoning.

\subsection{Reinforcement Fine-Tuning for MLLMs.}
Reinforcement Learning~(RL) has increasingly surpassed SFT for optimizing reasoning in MLLMs. 
The DeepSeek-R1~\cite{guo2025deepseek} paradigm demonstrated self-emerging reasoning via GRPO~\cite{shao2024deepseekmath}, inspiring multimodal extensions such as Visual-RFT~\cite{liu2025visual} and MM-Eureka~\cite{meng2025mm}. 
For video, Video-R1~\cite{feng2025video} proposes Temporal-Guided GRPO using frame-shuffling verification, SpaceR~\cite{ouyang2025spacer} introduces spatial map imagination, VideoChat-R1~\cite{li2025videochat} explores task-specific temporal rewards, and DeepVideo-R1~\cite{park2025deepvideo} addresses vanishing-advantage problems with difficulty-aware GRPO. 
In the egocentric domain, EgoThinker~\cite{pei2025egothinker} introduces an IoU reward for spatio-temporal grounding. However, IoU-based rewards evaluate spatial overlap at the output level without examining whether the model's reasoning correctly identifies entities, respects temporal boundaries, or maintains logical consistency. 
Our task-aware rewards address this by verifying the structured reasoning traces directly: checking entity grounding, temporal alignment, and task-adaptive logical constraints~(\eg, trajectory sequences, angular accuracy, duration arithmetic).

\section{EgoReasoner}
We detail our framework for enhancing the 4D reasoning capabilities of Qwen2.5-VL~\cite{wang2024qwen2}. 
\cref{sec: dataset_const} describes our automated metadata-driven data curation pipeline, and \cref{sec:rft} introduces our two-stage optimization: structured cold-start (SFT) followed by grounded reinforcement fine-tuning (RFT) with task-aware rewards.

\subsection{\label{sec: dataset_const}Metadata-Driven QA and CoT Generation}
As discussed in \cref{sec:intro}, 4D egocentric reasoning requires training data with precise spatial-temporal grounding, something that standard video captioning or manual annotation cannot efficiently provide. 
We address this by developing an automated pipeline that extracts and fuses multi-modal metadata from egocentric videos to synthesize high-fidelity QA pairs and structured CoT reasoning traces, as illustrated in \cref{fig:data}.

\noindent \textbf{Data source.}
We utilize egocentric videos from the kitchen-domain subset of the Ego-Exo4D dataset~\cite{grauman2024ego}. 
Our training corpus consists of 443 videos totaling 56 hours of footage, with an average video duration of 7–8 minutes. These videos provide the complex, long-horizon interactions necessary for training robust 4D reasoning models.

\noindent \textbf{Automatic Generation Pipeline.}
We develop a systematic pipeline to synthesize training data by extracting object-centric spatial and temporal information from raw video sources and narrations:
\begin{enumerate}[label=(\arabic*), leftmargin=*, nosep]
    \item \textbf{Spatial Grounding:} Similar to data preprocessing in ~\cite{ashutosh2025fiction}, we employ the Detic~\cite{zhou2022detecting} segmentation model to generate 2D object masks across video frames. 
    Utilizing the 2D-3D alignment from Ego-Exo4D's SLAM-calibrated cameras, we project these masks into a 3D point cloud, yielding precise 2D bounding boxes and 3D object localizations for every frame.
    \item \textbf{Temporal and Semantic Alignment:} We utilize Gemini~\cite{comanici2025gemini} to refine the accompanying textual narrations into structured action descriptions, each strictly anchored to its corresponding timestamp in the video.
    \item \textbf{Metadata Fusion:} We perform spatio-temporal matching to align refined narrations with the spatial grounding data. 
    The resulting \textbf{4D Descriptions} encapsulate semantic actions, involved entities (fixtures and objects), and their dynamic 2D/3D trajectories, as shown in \cref{fig:data}.
\end{enumerate}
This metadata provides the foundation for generating hallucination-free QA pairs and CoT reasoning traces that are grounded in the physical reality of each video.

\noindent \textbf{QA Generation.}
Using the synthesized \textbf{4D Descriptions}, we employ Gemini~\cite{comanici2025gemini} to generate complex, multi-hop QA pairs via in-context learning. 
Since raw egocentric videos often exceed 20 minutes, we apply a temporal windowing strategy, focusing generation on 2–5 minute clips to maintain reasoning density. 
To ensure precise grounding, the LLM undergoes a visual re-verification phase using sampled video frames, during which we introduce challenging distractors that are visually or temporally similar to the ground truth.
We further incorporate relative positional descriptors~(\eg, ``chopping board left of sink'' rather than generic ``countertop'') to force fine-grained object identification. 
Finally, we adopt a human-in-the-loop review to iteratively refine prompts and remove incorrect or low-quality QA pairs.

In total, we curate 4.3K QA pairs for Stage I supervised fine-tuning~(SFT) and 12K pairs for Stage II Reinforcement fine-tuning~(RFT), all with complete Chain-of-Thought reasoning traces.

\noindent \textbf{Structured CoT Synthesis.}
A key challenge is generating CoT traces that reflect the distinct reasoning structures required by each task type - the motivation behind our task-adaptive thinking templates (introduced in \cref{sec:intro}).
We design unique cognitive templates for each task category, established via human expertise and refined by Gemini~\cite{comanici2025gemini}. 
As illustrated in \cref{fig:think_temp}, these templates decompose 4D reasoning into structured sub-steps that explicitly \hlyellow{ground entities} 
and \hlred{spatial-temporal metadata}, such as clock-face angles and trajectory segments. 
By prompting Gemini~(the teacher model) with the 4D metadata and the final answer, it ``reasons backward'' to reconstruct the logical steps required to reach the conclusion from raw visual inputs. 
These synthesized traces serve as the primary supervision for both Stage I SFT and Stage II RFT. 
Detailed instruction prompts are provided in the Supplementary.

\subsection{\label{sec:rft}Two-Stage Optimization via Structured Reasoning and RFT}
As illustrated in \cref{fig:train_pipeline}, we adopt a two-stage strategy: Stage I establishes reasoning format and spatial-temporal priors via SFT, while Stage II enforces grounded, logically consistent reasoning via GRPO with task-aware rewards.

\noindent \textbf{Stage I: Structured Cold-Start~(SFT)}
In this initial stage, we perform supervised fine-tuning on our curated dataset. 
Unlike standard SFT that supervises only the final answer, we train the student model (Qwen2.5-VL~\cite{wang2024qwen2}) directly on the task-adaptive thinking templates described in \cref{fig:think_temp}.
The model learns to output detailed reasoning traces encapsulated within \texttt{<think> </think>} tags, followed by the final answer enclosed in \texttt{<answer> </answer>} tags, as shown in \cref{fig:train_pipeline}. 
This cold-start phase serves two purposes: it teaches the model the required logical format for each task type, and it elicits spatial-temporal priors~(\eg, clock-face angular reasoning, trajectory enumeration) that provide a stable foundation for subsequent reinforcement learning.

\noindent \textbf{Stage II: Grounded Reinforcement Fine-Tuning (RFT)}
While Stage I teaches the model how to structure its reasoning, supervised imitation cannot guarantee that the reasoning is factually grounded in the video — the model may produce well-formatted but incorrect traces. 
To enforce physical accuracy, we further optimize the model using Group Relative Policy Optimization~(GRPO)~\cite{shao2024deepseekmath}.

For each query $q$~(a video-question pair), the policy $\pi_{\theta}$ generates a group of $G$ reasoning trajectories $\{o_1, o_2, \dots, o_G\}$ through sampling. 
The policy update maximizes the expected advantage-weighted likelihood:

\begin{equation}
\max_{\pi_{\theta}} \mathbb{E}{o \sim \pi{\theta_{\text{old}}}} \left[ \sum_{i=1}^{G} \frac{\pi_{\theta}(o_i|q)}{\pi_{\theta_{\text{old}}}(o_i|q)} \cdot A_i - \beta D_{\text{KL}}(\pi_{\theta} | \pi_{\text{ref}}) \right],
\end{equation}
\noindent where $\pi_{\theta_{\text{old}}}$ is the policy from the previous iteration, $\beta$ is the coefficient for the KL divergence coefficient, and $\pi_{\text{ref}}$ is the reference model (the Stage I SFT model) used to prevent catastrophic forgetting or format drift. 
The normalized advantage $A_i$ for the $i$-th trajectory is:

\begin{equation}
    A_i = \frac{r_i - \text{mean}(\{r_1, r_2, \dots, r_G\})}{\text{std}(\{r_1, r_2, \dots, r_G\})}
\end{equation}

\noindent Here, $r_i$ is the reward score from our task-aware reward functions. 
This formulation reinforces reasoning steps that lead to physically grounded answers while penalizing paths that drift from the reference model.

\noindent \textbf{Task-Aware Reward Functions.}
The core innovation of Stage II is our suite of task-aware reward functions that provide granular feedback on the reasoning process, not just the final answer. 
We use regex-based parsing to extract key entities and metadata from the generated reasoning traces, comparing them against the ground-truth 4D Descriptions. 
This is made reliable by Stage I's structured cold-start, which ensures the model consistently produces well-formatted traces with predictable tag patterns that regex can robustly parse. 
The total reward $r$ is a weighted composite of four components:
\begin{equation}
    r = \alpha R_{acc} + \lambda R_{grd} + \gamma R_{log} + \delta R_{struct},
\end{equation}
where coefficients $\alpha, \lambda, \gamma, \text{ and } \delta$ are hyperparameters balancing task performance and reasoning quality. 
Each component targets a distinct aspect of reasoning correctness:

\begin{itemize}[leftmargin=*, nosep]
\item \textbf{Accuracy Reward ($R_{acc}$):} A binary reward (0 or 1) evaluating the correctness of the final answer choice (\eg, A–E) against the ground truth.
\item \textbf{Grounding Reward ($R_{grd}$):} This reward ensures the model's reasoning is anchored in the video's physical metadata, via two binary sub-rewards.
\begin{itemize}[leftmargin=1.5em, nosep]
\item \textit{Entity Grounding:} 
checks whether the regex-extracted entity in Step 0/1 (\eg, \hlyellow{\texttt{[Fixture Name]}}/\hlyellow{\texttt{[Object Name]}} in \cref{fig:think_temp}) matches the target entity in the metadata. 
\item \textit{Temporal Grounding:} 
uses a soft-matching window (\eg, $\pm 2.0s$):
If a predicted timestamp (\hlyellow{\texttt{<TIME>}}) falls within the threshold of the ground-truth event, it receives a full reward; otherwise, it is penalized.
\end{itemize}
\item \textbf{Logic Reward ($R_{log}$):} 
This reward evaluates task-adaptive internal consistency, varying between binary and continuous signals depending on the task. 
\begin{itemize}[leftmargin=1.5em, nosep]
\item \textit{Fixture Verification:} 
gives a binary reward if the model correctly identifies the associated \hlred{\texttt{[Fixture Name]}} in the grounding step.
\item \textit{Duration and Sequence Verification:} 
checks that calculated values, such as the duration of a static window (\hlred{$t_1$}) or the number of enumerated \hlred{\texttt{\{Event\}}}/\hlred{\texttt{\{Track\}}} entries, match the metadata.
\item \textit{Angular Accuracy:} computes a continuous reward based on circular distance $d$ on a 12-hour \hlred{clock-face}: $R_{log} = \max(0, 1 - d/6)$.
\end{itemize}
\item \textbf{Format Reward ($R_{struct}$):} A deterministic check ensuring the model adheres to the required \texttt{<think>} and \texttt{<answer>} tag structure.
\end{itemize}

This multi-faceted reward design directly addresses the challenges outlined in \cref{sec:intro}: the grounding rewards catch entity misidentification and temporal drift, the logic rewards enforce task-adaptive spatial-temporal constraints, and the format reward maintains the structured reasoning format established in Stage I.

\section{Experiments}
\label{sec:experiment}
\subsection{Setup}
We conduct all experiments using Qwen2.5-VL~\cite{wang2024qwen2} trained on data curated from Ego-Exo4D~\cite{grauman2024ego}. 
We evaluate on the HD-EPIC benchmark~\cite{perrett2025hd}, which comprises six tasks in 4D space across two categories: Egocentric 3D Perception (Fixture Interaction Counting, Fixture Location, Object Location) and Egocentric Object Motion (Object Movement Counting, Object Movement Itinerary, Stationary Object Localization). 
As illustrated in \cref{fig:task}, these tasks require structured spatial-temporal reasoning that goes well beyond simple object detection or action recognition. 
More benchmark evaluations are in the Supplementary.

\subsection{Implementation Details}
We perform temporal clipping based on question timestamps, reducing average video length from 15–30 minutes to 2–5 minutes. 
All baselines and our method use these synchronized clips for fair comparison. 
We initialize with Qwen2.5-VL-3B-Instruct~\cite{wang2024qwen2}, sampling up to 128 frames at 256×256 resolution, and train on 8 NVIDIA H100 GPUs (~24 hours total).

The training process has two sequential stages. 
In the first Supervised Fine-Tuning (SFT) stage, we train on our curated dataset of 4.3K QA pairs while freezing the vision encoder and performing full-parameter tuning on the projector and LLM backbone. 
We use a batch size of 16 and a learning rate of $5 \times 10^{-6}$ optimized with AdamW. 
The second stage utilizes Group Relative Policy Optimization (GRPO) to further align the model's multi-step reasoning chains with ground-truth spatial-temporal metadata. For this RL tuning phase, we set the global batch size to 256, the group size ($G$) to 5, and the KL divergence coefficient ($\beta$) to 0.04. 
The learning rate is $5 \times 10^{-6}$, and the maximum response length is limited to 4096 tokens.
Optimal reward weights for each task are determined via hyperparameter search, and the resulting task-adaptive configurations are subsequently integrated into the unified training stage. 
We put details in the Supplementary due to space limit.

\subsection{Main Results}
We evaluate our framework on different tasks from HD-EPIC benchmark and observe a clear evolutionary improvement in reasoning capabilities as we move from standard supervised learning to task-aware reinforcement learning. 
Our results are summarized in \cref{tab:quan_res} and \cref{tab:sep_train}.

\begin{table}[t]
    \centering
    \caption{\label{tab:quan_res} \textbf{Quantitative results on the HD-EPIC benchmark across six reasoning tasks.} We report multiple-choice accuracy for both closed/open-source baselines and our two-stage approach.}
    \resizebox{\linewidth}{!}{
    \begin{tabular}{l c | c | c c c | c c c}
        \toprule
        \multirow{3}{*}{\textbf{Models}} & \multirow{3}{*}{\textbf{Size}} & \multirow{3}{*}{\textbf{Avg.}} & \multicolumn{3}{c}{\textbf{3D Perception}} & \multicolumn{3}{c}{\textbf{Object Motion}}  
        \\
        \cmidrule(lr){4-6} \cmidrule(lr){7-9} 
        & & & \multicolumn{1}{c|}{\makecell{Fixture \\Interaction Count}} & \multicolumn{1}{c|}{\makecell{Fixture\\Location}} & \makecell{Object\\Location} & \multicolumn{1}{c|}{\makecell{Object \\Movement Count}} & \multicolumn{1}{c|}{\makecell{Object \\Movement Itinerary}} & \makecell{Stationary\\Object Localization} \\
        \midrule
        \rowB \textit{Sample Human Baseline} & -- & 91.1 & 89.5 & 95.6 & 96.3 & 87.6 & 93.2 & 84.3 \\
        \midrule
        \rowA \multicolumn{9}{l}{\textit{Closed-source Models}} \\
        Gemini 2.5 Pro~\cite{comanici2025gemini} & ${\sim}$128B & 49.2 & 36.3 & 40.4 & 63.6 & 45.5 & 60.0 & 49.5 \\
        \midrule
        \rowA \multicolumn{9}{l}{\textit{Open-source Models}} \\
        VideoLlama 2~\cite{cheng2024videollama} & 7B & 25.5 & 17.7 & 18.8 & 31.0 & 44.0 & 11.0 & 30.5 \\
        LongVA~\cite{zhang2024long}      & 7B & 28.1 & 32.3 & 26.6 & 41.2 & 34.5 & 10.2 & 23.5  \\
        LLaVA-Video~\cite{zhang2024video} & 7B & 20.9 & 16.3 & 21.8 & 30.6 & 20.0 & 9.8 & 27.0 \\
        VideoLlama 3~\cite{zhang2025videollama} & 7B & 26.0 & 27.7 & 22.4 & 35.6 & 30.5 & 18.2 & 21.5   \\
        EgoThinker~\cite{pei2025egothinker} & 7B & 27.4 & 28.3 & 20.0 & 37.4 & 30.0 & 25.0 & 23.5 \\
        VideoChat-R1~\cite{li2025videochat} & 7B & 19.2 & 17.0 & 19.2 & 22.8 & 21.5 & 13.4 & 21.5 \\
        Qwen-2.5-VL~\cite{wang2024qwen2} & 3B & 27.0 & 31.0 & 18.4 & 37.8 & 39.0 & 16.2 & 19.5  \\
        Qwen-2.5-VL~\cite{wang2024qwen2} & 7B & 25.7 & 22.7 & 22.6 & 34.0 & 33.0 & 20.2 & 22.0  \\
        \midrule
        \rowA \multicolumn{9}{l}{\textit{Ours}} \\
        Stage 1 - SFT & 3B & 32.2 & 29.3 & 29.2 & 42.0 & 44.0 & 28.0 & 20.5 \\
        Stage 2 - RFT & 3B & 37.5 & 32.7 & 30.6 & 50.4 & 59.5 & 30.5 & 21.0\\
        \bottomrule
    \end{tabular}
    }   
\end{table}

\noindent \textbf{Progressive Improvements with CoT and Task-Aware Rewards.}
As detailed in \cref{tab:sep_train}, we initially conducted experiments by training on individual tasks to isolate the impact of specific training components using the Qwen-2.5-VL-3B~\cite{wang2024qwen2} backbone. 
Incorporating our specialized thinking templates as Chain-of-Thought (CoT) reasoning traces during the SFT stage yielded an immediate performance boost in complex reasoning tasks. 
For instance, compared to standard SFT without CoT, accuracy for \textit{Fixture Interaction Counting} increases significantly from 24.3\% to 34.3\% (+10.0\%).

The transition to Reinforcement Learning tuning~(RFT), provides the most substantial gains when training tasks independently. 
By introducing task-aware reward functions, the model demonstrates markable improvements in temporal tracking, particularly in the \textit{Object Movement Itinerary} task, which improved from 26.2\% to 36.6\%. 
As shown in \cref{tab:sep_train}, RL tuning with task-aware rewards consistently outperforms tuning without such guidance. 
We observed that the \textit{Stationary Object Localization} task remains an outlier in performance. 
This is primarily due to the extreme length of the input videos~(typically 8--10 minutes), which presents a significant challenge for the model's temporal context window. 
This difficulty in digesting long-form video, as also noted in \cite{cai2024temporalbench}, makes it difficult for RFT to effectively correct the model's reasoning traces.

The task-aware rewards listed in \cref{tab:sep_train} represent a curated combination of different reward functions as detailed in \cref{sec:rft}. 
We performed ablation studies on these individual rewards to determine and select the optimal strategy for each specific reasoning task.
\begin{figure}[t]
    \centering
    \includegraphics[width=\linewidth]{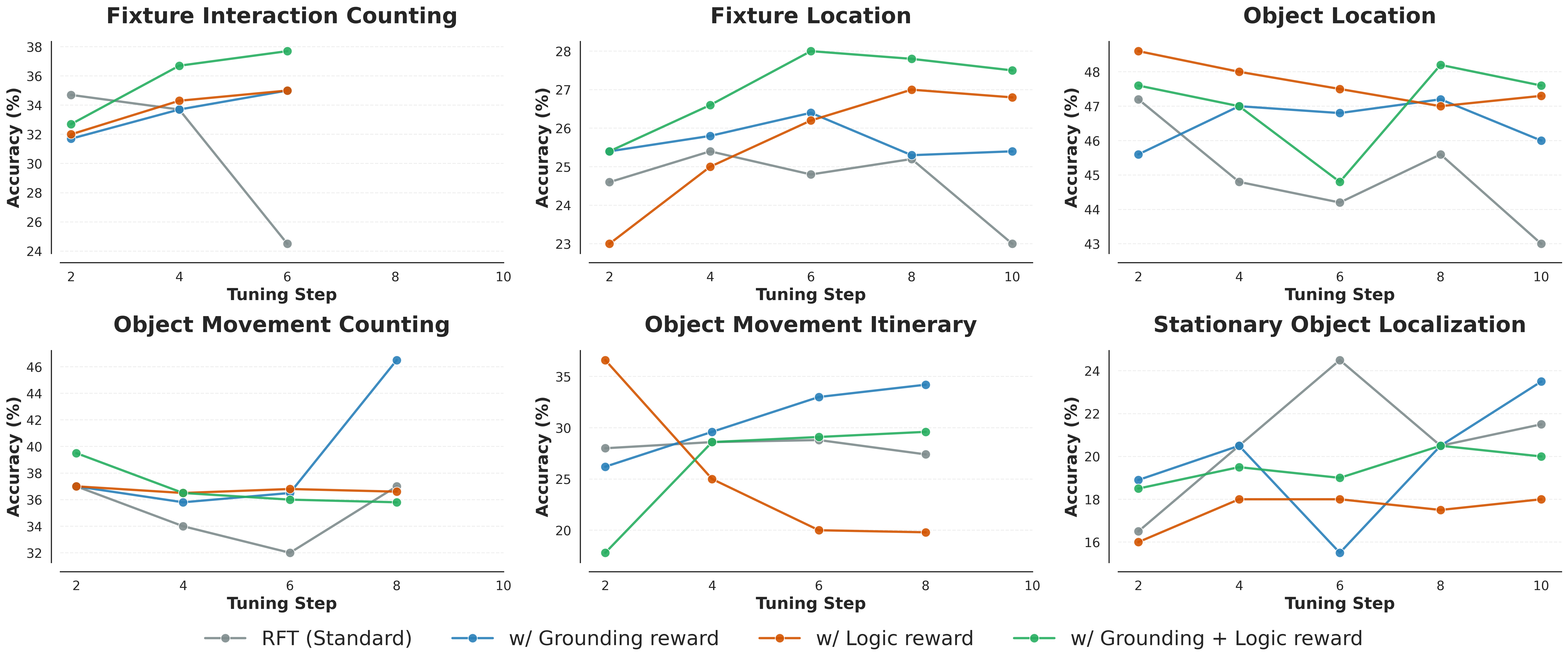}
    \caption{\label{fig:rl_step_perf}\textbf{Impact of task-aware rewards during RFT.} We compare \textcolor[HTML]{7f8c8d}{standard RFT}~(without any task-aware rewards) against variants using \textcolor[HTML]{2980b9}{Grounding}, \textcolor[HTML]{d35400}{Logic}, and \textcolor[HTML]{27ae60}{Combined} rewards across six tasks. } 
\end{figure}

\noindent \textbf{Unified Training and Task Synergy.}
We consolidate all tasks into a unified pipeline, incorporating the optimal reward strategies identified during separate training.
As shown in \cref{tab:quan_res}, multi-task training promotes cross-task synergy in both stages.
These tasks are exceptionally challenging---Qwen2.5-VL-7B (25.7\%) does not outperform the 3B model (27.0\%), and both trail the human baseline (91.1\%) by a wide margin.
We therefore adopt the 3B model for our primary experiments.

Our unified Stage~2 model achieves 37.5\% average accuracy (+11.8 over the 7B baseline), with notable gains on \textit{Object Movement Counting}
(59.5\%, +14.0 over the strongest baseline) and \textit{Object Location}
(50.4\%).
These results demonstrate that RFT strengthens temporal tracking without
degrading spatial perception.
\textit{Stationary Object Localization} remains an outlier---its 8--10 minute
videos far exceed the 3--5 minute clips of other tasks, limiting RFT's
ability to correct reasoning at this temporal
scale~\cite{cai2024temporalbench}.
We also observe that 3D Perception tasks benefit more from SFT-stage synergy, while Object Motion tasks gain primarily from RFT, suggesting that SFT establishes spatial-semantic foundations which RFT then refines for temporal reasoning.

\begin{table}[t]
\centering
\small
\caption{\label{tab:sep_train} \textbf{Ablation of different tasks with different settings training separately.} We use Qwen2.5-VL-3B model for experiments. Bold indicates best; \underline{underline} indicates second best. \textbf{FIC}: Fixture Interaction Count, \textbf{FL}: Fixture Location, \textbf{OL}: Object Location, \textbf{OMC}: Object Movement Count, \textbf{OMI}: Object Movement Itinerary, \textbf{SOL}: Stationary Object Localization.}
\resizebox{0.8 \linewidth}{!}{
\begin{tabular}{l | cc | ccc | ccc | c}
\toprule
\multirow{2}{*}{\textbf{Setting}} & \multicolumn{2}{c|}{\textbf{Variant}} & \multicolumn{3}{c|}{\textbf{3D Perception}} & \multicolumn{3}{c|}{\textbf{Object Motion}} & \multirow{2}{*}{\textbf{Avg.}}\\
\cmidrule(lr){2-3} \cmidrule(lr){4-6} \cmidrule(lr){7-9}
 & CoT & \makecell{Task-Aware Reward} & FIC & FL & OL & OMC & OMI & SOL & \\
\midrule
Baseline & - & - & 31.0 & 18.4 & 37.8 & 39.0 & 16.2 & 19.5 & 27.0\\
\midrule
\multirow{2}{*}{+SFT} & - & - & 24.3 & 22.6 & 38.4 & \underline{42.5} & 21.6 & 20.0 & 28.2 \\
     & \checkmark & - & \underline{34.3} & \underline{25.2} & \underline{44.8} & 41.0 & \underline{28.0} & 18.0 & \underline{31.9} \\
\midrule
\multirow{2}{*}{+RFT} & \checkmark & - & 33.7 & 24.8 & 40.6 & 37.0 & 26.2 & \textbf{24.5} & 31.1 \\
 & \checkmark & \checkmark & \textbf{37.7} & \textbf{28.0} & \textbf{48.2} & \textbf{46.5} & \textbf{36.6} & \underline{23.5} & \textbf{36.8}\\
\bottomrule
\end{tabular}
}
\end{table}

\begin{figure}[H]
    \centering
    \includegraphics[width=\linewidth]{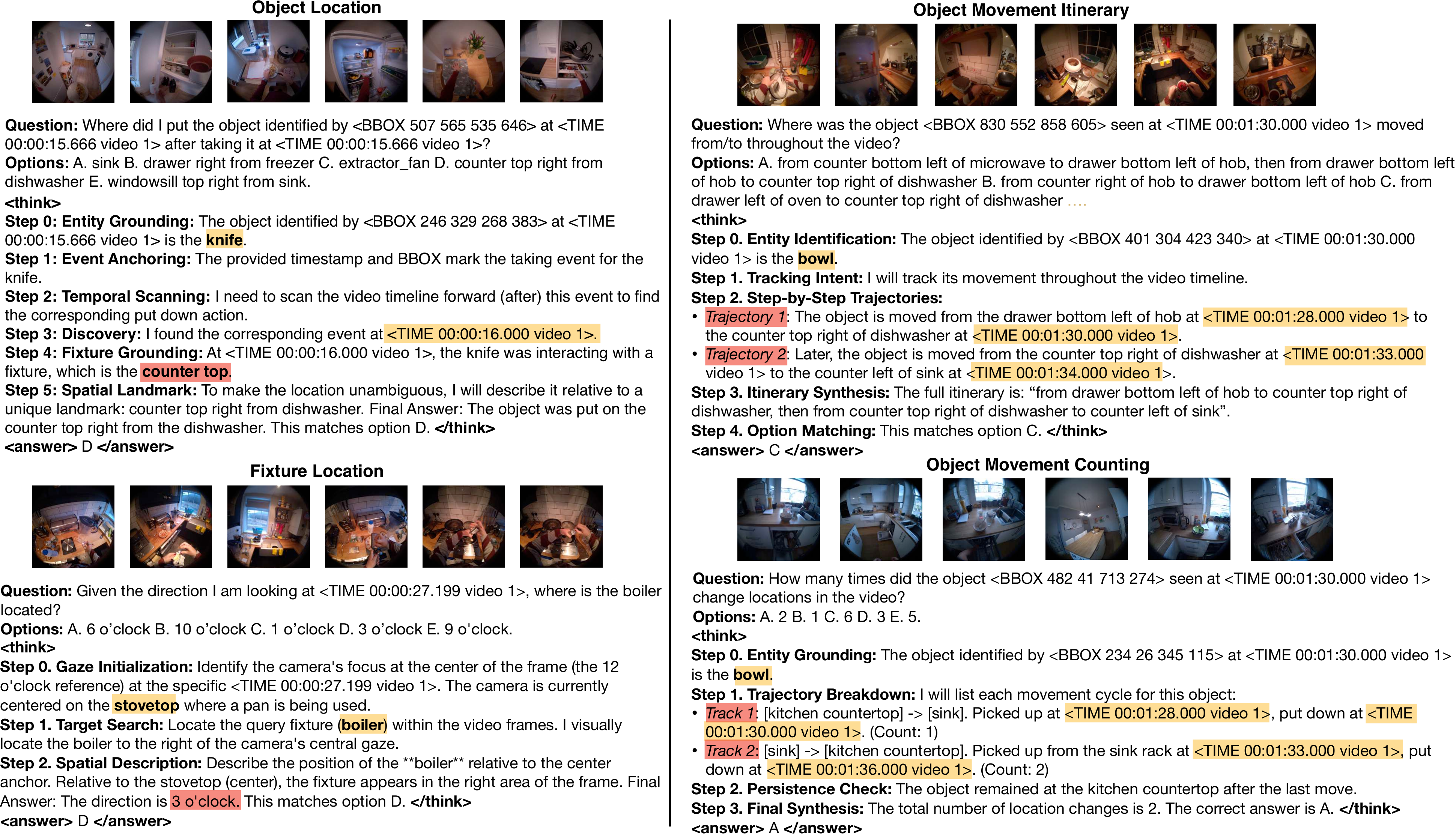}
    \caption{\label{fig:vis_res} \textbf{Qualitative results of different tasks on the HD-EPIC benchmark.} We demonstrate how \model{} successfully handles complex spatial and temporal queries through structured thinking traces. \hlyellow{Yellow highlights} indicate grounded entity names (fixtures and objects) / timestamps, while \hlred{red highlights} denote specific spatial-temporal metadata. Some options are omitted due to space limit.} 
\end{figure}
\noindent \textbf{Qualitative Analysis}
We provide qualitative examples across different task categories in ~\cref{fig:vis_res}~(more in Supplementary). 
\model{} effectively utilizes the structured thinking traces developed in Stage I and refined in Stage II to decompose complex queries into verifiable steps. 
In the Object Location example, the model successfully grounds the target object (``knife'') at its starting timestamp before correctly identifying its final destination (``counter top'') by scanning the temporal sequence. 
Similarly, for Object Motion tasks like Object Movement Itinerary, the model demonstrates precise temporal tracking by breaking down the object's path into distinct trajectory segments (Track 1 \& 2), ensuring the final synthesis remains physically plausible. 
These results highlight the model's ability to maintain high-fidelity perception and logical consistency, under the significant ego-motion and dynamic occlusion typical of first-person video.

\subsection{Ablation Study}
To evaluate the impact of our reward design, we track the accuracy of each task throughout the RFT process with different reward functions in \cref{fig:rl_step_perf}.
\textcolor[HTML]{7f8c8d}{Standard RFT}, which lacks task-aware rewards, exhibits significant instability and performance degradation across most categories as tuning steps increase. 
This is particularly evident in \textit{Fixture Interaction Counting} and \textit{Fixture Location}, where accuracy drops sharply after initial steps, suggesting that vanilla RFT struggles to maintain spatial-semantic alignment in high-density egocentric environments.

The introduction of specialized rewards provides critical stabilization. 
The \textcolor[HTML]{27ae60}{Combined} reward strategy generally achieves superior stability and higher peak performance. 
We observe distinct roles for each component:
\begin{itemize}
\item \textbf{Grounding Reward:} As shown in the \textit{Object Movement Counting} and \textit{Itinerary} tasks, the \textcolor[HTML]{2980b9}{Grounding} reward is essential for aligning temporal reasoning with physical object trajectories. 
In \textit{Object Movement Counting}, it enables a continuous upward performance trend, unlike the stagnant or declining trends of other variants.
\item \textbf{Logic Reward:} The \textcolor[HTML]{d35400}{Logic} reward proves vital for tasks requiring structured spatial reasoning, such as \textit{Object Location}. It often provide a higher starting accuracy and stabilizes the reasoning chain, preventing the catastrophic forgetting observed in standard RFT.
\end{itemize}
The \textcolor[HTML]{27ae60}{Combined} strategy effectively elicits synergy between these components, leading to the most robust results in most tasks. 
The exception remains \textit{Stationary Object Localization}, where all variants struggle to maintain consistent gains due to the extreme temporal context requirements.

\section{Conclusion}

We introduce \model{}, a framework that addresses the challenges of 4D egocentric reasoning through two key innovations: task-adaptive thinking templates that decompose each reasoning task into grounded sub-steps, and task-aware rewards for GRPO that enforce entity grounding, temporal alignment, and logical consistency beyond what standard SFT or coarse-reward RL can achieve. 
Supported by an automated metadata-driven pipeline that curate high-fidelity training data from SLAM-calibrated 3D detections and refined narrations, our framework significantly outperforms general-purpose MLLMs on several reasoning tasks. Our results demonstrate that task-adaptive structured reasoning combined with fine-grained reinforcement learning is a scalable and effective approach for 4D egocentric scene understanding. 
Our work paves the way for metadata-grounded reasoning in embodied AI agents that need to navigate and reason within complex, dynamic environments.

\bibliographystyle{splncs04}
\bibliography{main}

\begin{thebibliography}{10}
\providecommand{\url}[1]{\texttt{#1}}
\providecommand{\urlprefix}{URL }
\providecommand{\doi}[1]{https://doi.org/#1}

\bibitem{ashutosh2025fiction}
Ashutosh, K., Pavlakos, G., Grauman, K.: Fiction: 4d future interaction prediction from video. In: Proceedings of the Computer Vision and Pattern Recognition Conference (2025)

\bibitem{cai2024temporalbench}
Cai, M., Tan, R., Zhang, J., Zou, B., Zhang, K., Yao, F., Zhu, F., Gu, J., Zhong, Y., Shang, Y., et~al.: Temporalbench: Benchmarking fine-grained temporal understanding for multimodal video models. arXiv preprint arXiv:2410.10818  (2024)

\bibitem{cheng2024spatialrgpt}
Cheng, A.C., Yin, H., Fu, Y., Guo, Q., Yang, R., Kautz, J., Wang, X., Liu, S.: Spatialrgpt: Grounded spatial reasoning in vision-language models. Advances in Neural Information Processing Systems  \textbf{37},  135062--135093 (2024)

\bibitem{cheng2024compressed}
Cheng, J., Van~Durme, B.: Compressed chain of thought: Efficient reasoning through dense representations. arXiv preprint arXiv:2412.13171  (2024)

\bibitem{cheng2024videollama}
Cheng, Z., Leng, S., Zhang, H., Xin, Y., Li, X., Chen, G., Zhu, Y., Zhang, W., Luo, Z., Zhao, D., et~al.: Videollama 2: Advancing spatial-temporal modeling and audio understanding in video-llms. arXiv preprint arXiv:2406.07476  (2024)

\bibitem{comanici2025gemini}
Comanici, G., Bieber, E., Schaekermann, M., Pasupat, I., Sachdeva, N., Dhillon, I., Blistein, M., Ram, O., Zhang, D., Rosen, E., et~al.: Gemini 2.5: Pushing the frontier with advanced reasoning, multimodality, long context, and next generation agentic capabilities. arXiv preprint arXiv:2507.06261  (2025)

\bibitem{fei2024video}
Fei, H., Wu, S., Ji, W., Zhang, H., Zhang, M., Lee, M.L., Hsu, W.: Video-of-thought: Step-by-step video reasoning from perception to cognition. arXiv preprint arXiv:2501.03230  (2024)

\bibitem{feng2025video}
Feng, K., Gong, K., Li, B., Guo, Z., Wang, Y., Peng, T., Wu, J., Zhang, X., Wang, B., Yue, X.: Video-r1: Reinforcing video reasoning in mllms. arXiv preprint arXiv:2503.21776  (2025)

\bibitem{ghazanfari2025chain}
Ghazanfari, S., Croce, F., Flammarion, N., Krishnamurthy, P., Khorrami, F., Garg, S.: Chain-of-frames: Advancing video understanding in multimodal llms via frame-aware reasoning. arXiv preprint arXiv:2506.00318  (2025)

\bibitem{grauman2022ego4d}
Grauman, K., Westbury, A., Byrne, E., Chavis, Z., Furnari, A., Girdhar, R., Hamburger, J., Jiang, H., Liu, M., Liu, X., et~al.: Ego4d: Around the world in 3,000 hours of egocentric video. In: Proceedings of the IEEE/CVF conference on computer vision and pattern recognition. pp. 18995--19012 (2022)

\bibitem{grauman2024ego}
Grauman, K., Westbury, A., Torresani, L., Kitani, K., Malik, J., Afouras, T., Ashutosh, K., Baiyya, V., Bansal, S., Boote, B., et~al.: Ego-exo4d: Understanding skilled human activity from first-and third-person perspectives. In: Proceedings of the IEEE/CVF Conference on Computer Vision and Pattern Recognition. pp. 19383--19400 (2024)

\bibitem{guo2025deepseek}
Guo, D., Yang, D., Zhang, H., Song, J., Wang, P., Zhu, Q., Xu, R., Zhang, R., Ma, S., Bi, X., et~al.: Deepseek-r1: Incentivizing reasoning capability in llms via reinforcement learning. arXiv preprint arXiv:2501.12948  (2025)

\bibitem{kazakos2019epic}
Kazakos, E., Nagrani, A., Zisserman, A., Damen, D.: Epic-fusion: Audio-visual temporal binding for egocentric action recognition. In: ICCV (2019)

\bibitem{li2025think}
Li, M., Zhong, J., Zhao, S., Lai, Y., Zhang, H., Zhu, W.B., Zhang, K.: To think or not to think: A study of thinking in rule-based visual reinforcement fine-tuning. In: The Thirty-ninth Annual Conference on Neural Information Processing Systems (2025)

\bibitem{li2025videochat}
Li, X., Yan, Z., Meng, D., Dong, L., Zeng, X., He, Y., Wang, Y., Qiao, Y., Wang, Y., Wang, L.: Videochat-r1: Enhancing spatio-temporal perception via reinforcement fine-tuning. arXiv preprint arXiv:2504.06958  (2025)

\bibitem{liu2025visual}
Liu, Z., Sun, Z., Zang, Y., Dong, X., Cao, Y., Duan, H., Lin, D., Wang, J.: Visual-rft: Visual reinforcement fine-tuning. In: Proceedings of the IEEE/CVF International Conference on Computer Vision. pp. 2034--2044 (2025)

\bibitem{ma2025spatialreasoner}
Ma, W., Chou, Y.C., Liu, Q., Wang, X., de~Melo, C., Xie, J., Yuille, A.: Spatialreasoner: Towards explicit and generalizable 3d spatial reasoning. arXiv preprint arXiv:2504.20024  (2025)

\bibitem{mangalam2023egoschema}
Mangalam, K., Akshulakov, R., Malik, J.: Egoschema: A diagnostic benchmark for very long-form video language understanding. Advances in Neural Information Processing Systems  \textbf{36},  46212--46244 (2023)

\bibitem{meng2025mm}
Meng, F., Du, L., Liu, Z., Zhou, Z., Lu, Q., Fu, D., Han, T., Shi, B., Wang, W., He, J., et~al.: Mm-eureka: Exploring the frontiers of multimodal reasoning with rule-based reinforcement learning. arXiv preprint arXiv:2503.07365  (2025)

\bibitem{ouyang2025spacer}
Ouyang, K., Liu, Y., Wu, H., Liu, Y., Zhou, H., Zhou, J., Meng, F., Sun, X.: Spacer: Reinforcing mllms in video spatial reasoning. arXiv preprint arXiv:2504.01805  (2025)

\bibitem{park2025deepvideo}
Park, J., Na, J., Kim, J., Kim, H.J.: Deepvideo-r1: Video reinforcement fine-tuning via difficulty-aware regressive grpo. arXiv preprint arXiv:2506.07464  (2025)

\bibitem{pei2025egothinker}
Pei, B., Huang, Y., Xu, J., He, Y., Chen, G., Wu, F., Qiao, Y., Pang, J.: Egothinker: Unveiling egocentric reasoning with spatio-temporal cot. arXiv preprint arXiv:2510.23569  (2025)

\bibitem{perrett2025hd}
Perrett, T., Darkhalil, A., Sinha, S., Emara, O., Pollard, S., Parida, K.K., Liu, K., Gatti, P., Bansal, S., Flanagan, K., et~al.: Hd-epic: A highly-detailed egocentric video dataset. In: Proceedings of the Computer Vision and Pattern Recognition Conference. pp. 23901--23913 (2025)

\bibitem{plizzari2022e2}
Plizzari, C., Planamente, M., Goletto, G., Cannici, M., Gusso, E., Matteucci, M., Caputo, B.: E2 (go) motion: Motion augmented event stream for egocentric action recognition. In: CVPR (2022)

\bibitem{aria}
{Project Aria Team}: {Aria Gen2: Bringing the Human Perspective to AI.} {\url{https://www.projectaria.com/}} (2025)

\bibitem{astra}
{Project Astra Team}: {Exploring breakthrough capabilities for Google products — on the way to building a universal AI assistant.} {\url{https://deepmind.google/models/project-astra/}} (2025)

\bibitem{shao2024visual}
Shao, H., Qian, S., Xiao, H., Song, G., Zong, Z., Wang, L., Liu, Y., Li, H.: Visual cot: Unleashing chain-of-thought reasoning in multi-modal language models. CoRR  (2024)

\bibitem{shao2024deepseekmath}
Shao, Z., Wang, P., Zhu, Q., Xu, R., Song, J., Bi, X., Zhang, H., Zhang, M., Li, Y., Wu, Y., et~al.: Deepseekmath: Pushing the limits of mathematical reasoning in open language models. arXiv preprint arXiv:2402.03300  (2024)

\bibitem{sudhakaran2019lsta}
Sudhakaran, S., Escalera, S., Lanz, O.: Lsta: Long short-term attention for egocentric action recognition. In: CVPR (2019)

\bibitem{tian2025ego}
Tian, S., Wang, R., Guo, H., Wu, P., Dong, Y., Wang, X., Yang, J., Zhang, H., Zhu, H., Liu, Z.: Ego-r1: Chain-of-tool-thought for ultra-long egocentric video reasoning. arXiv preprint arXiv:2506.13654  (2025)

\bibitem{wang2025think}
Wang, J., Lin, K.Q., Cheng, J., Shou, M.Z.: Think or not? selective reasoning via reinforcement learning for vision-language models. arXiv preprint arXiv:2505.16854  (2025)

\bibitem{wang2024qwen2}
Wang, P., Bai, S., Tan, S., Wang, S., Fan, Z., Bai, J., Chen, K., Liu, X., Wang, J., Ge, W., et~al.: Qwen2-vl: Enhancing vision-language model's perception of the world at any resolution. arXiv preprint arXiv:2409.12191  (2024)

\bibitem{wang2025videorft}
Wang, Q., Yu, Y., Yuan, Y., Mao, R., Zhou, T.: Videorft: Incentivizing video reasoning capability in mllms via reinforced fine-tuning. arXiv preprint arXiv:2505.12434  (2025)

\bibitem{wang2020symbiotic}
Wang, X., Zhu, L., Wu, Y., Yang, Y.: Symbiotic attention for egocentric action recognition with object-centric alignment. IEEE transactions on pattern analysis and machine intelligence  (2020)

\bibitem{wei2022chain}
Wei, J., Wang, X., Schuurmans, D., Bosma, M., Xia, F., Chi, E., Le, Q.V., Zhou, D., et~al.: Chain-of-thought prompting elicits reasoning in large language models. Advances in neural information processing systems  \textbf{35},  24824--24837 (2022)

\bibitem{wu2026st}
Wu, P., Liu, Y., Liu, M., Shen, J.: St-think: How multimodal large language models reason about 4d worlds from ego-centric videos. In: Proceedings of the IEEE/CVF Winter Conference on Applications of Computer Vision. pp. 5174--5183 (2026)

\bibitem{xu2025llava}
Xu, G., Jin, P., Wu, Z., Li, H., Song, Y., Sun, L., Yuan, L.: Llava-cot: Let vision language models reason step-by-step. In: Proceedings of the IEEE/CVF International Conference on Computer Vision. pp. 2087--2098 (2025)

\bibitem{ye2024mm}
Ye, H., Zhang, H., Daxberger, E., Chen, L., Lin, Z., Li, Y., Zhang, B., You, H., Xu, D., Gan, Z., et~al.: Mm-ego: Towards building egocentric multimodal llms for video qa. arXiv preprint arXiv:2410.07177  (2024)

\bibitem{zhang2025videollama}
Zhang, B., Li, K., Cheng, Z., Hu, Z., Yuan, Y., Chen, G., Leng, S., Jiang, Y., Zhang, H., Li, X., et~al.: Videollama 3: Frontier multimodal foundation models for image and video understanding. arXiv preprint arXiv:2501.13106  (2025)

\bibitem{zhang2025exo2ego}
Zhang, H., Chu, Q., Liu, M., Wang, Y., Wen, B., Yang, F., Gao, T., Zhang, D., Wang, Y., Nie, L.: Exo2ego: Exocentric knowledge guided mllm for egocentric video understanding. arXiv preprint arXiv:2503.09143  (2025)

\bibitem{zhang2024long}
Zhang, P., Zhang, K., Li, B., Zeng, G., Yang, J., Zhang, Y., Wang, Z., Tan, H., Li, C., Liu, Z.: Long context transfer from language to vision. arXiv preprint arXiv:2406.16852  (2024)

\bibitem{zhang2024video}
Zhang, Y., Wu, J., Li, W., Li, B., Ma, Z., Liu, Z., Li, C.: Video instruction tuning with synthetic data. arXiv preprint arXiv:2410.02713  (2024)

\bibitem{zhang2023multimodal}
Zhang, Z., Zhang, A., Li, M., Zhao, H., Karypis, G., Smola, A.: Multimodal chain-of-thought reasoning in language models. arXiv preprint arXiv:2302.00923  (2023)

\bibitem{zhou2022detecting}
Zhou, X., Girdhar, R., Joulin, A., Kr{\"a}henb{\"u}hl, P., Misra, I.: Detecting twenty-thousand classes using image-level supervision. In: ECCV (2022)

\bibitem{zhu2025struct2d}
Zhu, F., Wang, H., Xie, Y., Gu, J., Ding, T., Yang, J., Jiang, H.: Struct2d: A perception-guided framework for spatial reasoning in large multimodal models. arXiv preprint arXiv:2506.04220  (2025)

\end{thebibliography}
\end{document}